\pdfoutput=1

\documentclass[11pt]{article}

\usepackage[final]{acl}

\usepackage{times}
\usepackage{latexsym}

\usepackage[T1]{fontenc}

\usepackage[utf8]{inputenc}

\usepackage{microtype}

\usepackage{inconsolata}

\usepackage{graphicx}

\usepackage{subfig}
\usepackage{multicol}
\usepackage{booktabs, multirow} 
\usepackage{soul}
\usepackage{enumitem}
\usepackage{todonotes}
\usepackage{hyperref}
\usepackage{xcolor}
\usepackage{amsmath}
\usepackage{colortbl}
\usepackage{algorithm}
\usepackage{algorithmicx}
\usepackage{algpseudocode}
\usepackage{amsmath}
\usepackage{amssymb}
\usepackage{pgfplots}
\pgfplotsset{compat=1.17}
\usepackage{tikz}
\usepackage{xcolor}
\usepackage{booktabs}

\usepackage{array}    
\usepackage{rotating}  
\usepackage{amssymb}   

\usepackage{booktabs}
\usepackage{makecell}
\usepackage{xcolor}
\usepackage{siunitx}
\usepackage{multirow}
\usepackage{rotating}
\usepackage{booktabs}
\usepackage{multirow}
\usepackage{colortbl}
\usepackage{arydshln}
\usepackage{xcolor}
\usepackage{rotating}
\usepackage{listings}

\usetikzlibrary{shapes,arrows,positioning,shadows.blur}

\lstdefinelanguage{json}{
    basicstyle=\ttfamily\footnotesize, 
    numbers=none,
    stepnumber=1,
    showstringspaces=false,
    breaklines=true,
    frame=single,
    backgroundcolor=\color{gray!10},
    keywordstyle=\bfseries\color{blue},
    stringstyle=\color{red}
}

\usepackage{changepage,threeparttable} 
\usepackage{arydshln}

\newcommand{\model}{\texttt{HYPE}}
\definecolor{TableShade}{HTML}{F5F5F5} 
\title{Lifelong Model Editing with Graph-Based External Memory}

\author{Yash Kumar Atri \\
  University of Virginia\\
  \texttt{atri@virginia.edu} \\\And
  Ahmed Alaa \\
  UC Berkeley \& UCSF\\
  \texttt{amalaa@berkeley.edu} \\\And 
  Tom Hartvigsen \\
  University of Virginia\\
  \texttt{hartvigsen@virginia.edu}
}

\begin{document}
\maketitle
\begin{abstract}




Large language models (LLMs) have revolutionized natural language processing, yet their practical utility is often limited by persistent issues of hallucinations and outdated parametric knowledge. Although post-training model editing offers a pathway for dynamic updates, existing methods frequently suffer from overfitting and catastrophic forgetting. To tackle these challenges, we propose a novel framework that leverages hyperbolic geometry and graph neural networks for precise and stable model edits. We introduce \model\footnote{\model\ (\textbf{HY}perbolic \textbf{P}arameter \textbf{E}diting)}, which comprises three key components: (i) Hyperbolic Graph Construction, which uses Poincaré embeddings to represent knowledge triples in hyperbolic space, preserving hierarchical relationships and preventing unintended side effects by ensuring that edits to parent concepts do not inadvertently affect child concepts; (ii) Möbius-Transformed Updates, which apply hyperbolic addition to propagate edits while maintaining structural consistency within the hyperbolic manifold, unlike conventional Euclidean updates that distort relational distances; and (iii) Dual Stabilization, which combines gradient masking and periodic GNN parameter resetting to prevent catastrophic forgetting by focusing updates on critical parameters and preserving long-term knowledge. Experiments on CounterFact, CounterFact+, and MQuAKE with GPT-J and GPT2-XL demonstrate that \model\ significantly enhances edit stability, factual accuracy, and multi-hop reasoning.

\end{abstract}

\section{Introduction}

Large language models (LLMs) such as GPT-3 \cite{NEURIPS2020_1457c0d6}, PaLM \cite{10.5555/3648699.3648939}, and LLaMA \cite{llama} have revolutionized natural language processing (NLP), enabling unprecedented capabilities in text generation \cite{li-etal-2024-improving-faithfulness}, reasoning \cite{10.5555/3648699.3648939}, and contextual understanding \cite{zhao-etal-2024-enhancing}. These models underpin a wide range of applications, from conversational agents \cite{liu2024llmconversationalagentmemory} to knowledge-intensive tasks like question answering \cite{phukan-etal-2024-peering, shah-etal-2024-improving, gao-etal-2024-two} and summarization \cite{atri2023exploitingrepresentationbiasdata, 10261260, dey-etal-2020-corpora, atri-etal-2023-promoting}. However, their reliance on static, pre-trained parametric knowledge renders them prone to generating factual inaccuracies \cite{wang2024factualitylargelanguagemodels}, hallucinations \cite{10.1145/3571730}, and outdated information—critical limitations in real-world deployments where accuracy and timeliness are paramount \cite{10.5555/3540261.3542508, atri2025continuallyselfimprovinglanguagemodels}. While fine-tuning on updated data can mitigate these issues, the computational and data demands of retraining billion-parameter models \cite{kaplan2020scalinglawsneurallanguage} render this approach impractical for real-time knowledge updates. Instead, post-training model editing has emerged as a promising alternative, enabling targeted modifications to a model’s parametric knowledge without full retraining \cite{MEND}.

Existing model editing methods, such as ROME \cite{ROME} and MEMIT \cite{meng2022memit}, rely on Euclidean geometry to modify 
specific facts. While these methods achieve localized edits, they struggle with hierarchical relationships due to Euclidean space’s flat structure. For example, ROME enforces memorization via equality constraints, while MEMIT uses least-squares optimization, both of which fail to preserve relational depth \cite{yang2024hyperbolicfinetuninglargelanguage}. This leads to \textit{geometric mismatch}, where edits distort semantically related knowledge \cite{NIPS2017_59dfa2df}. Additionally, Euclidean updates in overparameterized models like GPT-2 \cite{gpt2-xl} cause \textit{update instability}, propagating unintended changes through dense connections \cite{kaplan2020scalinglawsneurallanguage}. Finally, these methods suffer from \textit{contextual fragility}, lacking mechanisms to maintain global coherence during edits \cite{zhong-etal-2023-mquake}.

Hyperbolic geometry, in contrast, naturally encodes hierarchical relationships through its exponential growth property, making it well-suited for modeling linguistic knowledge \cite{NIPS2017_59dfa2df}. Unlike Euclidean spaces, where distances are linear, hyperbolic space expands exponentially, allowing child nodes to be placed far from parent nodes while preserving their relational proximity \cite{10.5555/3327345.3327440}. This property enables precise parameter updates that do not propagate unintended side effects. For instance, modifying a parent concept in hyperbolic space affects only its immediate vicinity, preserving the stability of its hierarchical descendants. Furthermore, hyperbolic operations like Möbius addition \cite{ungar2013mobiustransformationeinstenvelocity} ensure updates remain on the manifold, avoiding the geometric distortion caused by Euclidean vector addition.

Motivated by these insights and the inherent advantages of hyperbolic geometry, we propose \textbf{HYPE} (\textbf{HY}perbolic \textbf{P}arameter \textbf{E}diting), a novel framework that leverages hyperbolic geometry \cite{10.5555/3327345.3327440} for model editing through three core components. First, Hyperbolic Graph Construction projects knowledge graph into a learnable curvature space using Poincaré embeddings \cite{NIPS2017_59dfa2df}, preserving hierarchical relationships—such as parent-child and whole-part associations—via exponential mapping and hyperbolic distances. Second, Möbius-Transformed \cite{ungar2013mobiustransformationeinstenvelocity} Updates adjust model parameters using hyperbolic addition, a geometric operation that preserves the hierarchical structure of the data. Unlike Euclidean vector addition, which can distort relationships in hierarchical data, Möbius addition ensures that edits remain within the hyperbolic manifold. This prevents unintended side effects by maintaining the relative distances between related concepts. Third, {Dual Stabilization} combines gradient-based sparsification \cite{frankle2019lotterytickethypothesisfinding}, which masks negligible gradient updates, with periodic resetting of graph network parameters to prevent the overwriting of critical pre-existing knowledge during the integration of new edits. By unifying hyperbolic embeddings, Möbius-Transformed updates, and robust stabilization techniques, \model\ achieves precise and resilient model edits.

\noindent We summarize our contributions as follows:
\begin{enumerate}[leftmargin=*,noitemsep,topsep=0pt]
    \item \model\ is the first approach to integrate hyperbolic geometry with graph-based model editing, leveraging Poincaré embeddings to project knowledge graphs into a learnable curvature space that preserves complex hierarchical relationships during model updates.
    
    \item \model\ introduces a novel editing mechanism that applies Möbius transformations for parameter updates, ensuring that edits remain consistent with the hyperbolic manifold and preserving the inherent hierarchical structure—thereby preventing the distortions common with Euclidean vector addition.
    
    \item We validate \model\ across three widely benchmarked datasets—CounterFact \cite{ROME}, CounterFact+ \cite{yao-etal-2023-editing}, and MQuAKE \cite{zhong-etal-2023-mquake}—and two popular LLMs—GPT-J \cite{gpt-j}, and GPT2-XL \cite{gpt2-xl}—demonstrating superior factual accuracy, parameter efficiency, and edit stability compared to existing model editing methods.
\end{enumerate}
Code available at: \url{https://github.com/yashkumaratri/HYPE}
\section{Related Work}
\label{sec:related}

\textbf{Post-Training Model Editing:}  
Model editing techniques broadly fall into parameter-modifying and parameter-preserving approaches. Parameter-modifying methods, such as locate-then-edit strategies \cite{ROME, meng2022memit, grace, kolbeinsson2025composable} and meta-learning frameworks \cite{MEND, KE}, adjust model parameters via hypernetworks or rank-one weight updates. \citet{GLAME} enhances specificity using knowledge graphs, but these methods typically treat parameter changes as isolated scalar operations \citep{PRUNE, RECT}, disregarding structured knowledge representations in LLMs \citep{nanda-etal-2023-emergent}. This often leads to unintended side effects \citep{side_effects} and poor support for compositional updates \citep{kolbeinsson2025composable}.  

Parameter-preserving approaches avoid direct weight updates by leveraging external memories or prompts \cite{SERAC, T-patcher, MemPrompt, IKE}, while continual learning methods \cite{OWM, OGD} use orthogonal projections to mitigate catastrophic forgetting. However, these strategies require architectural modifications or fail to maintain long-term coherence in factual updates.

 \begin{figure*}[!t] 
    \centering
    \scalebox{1}{
    \includegraphics[width=1\textwidth]{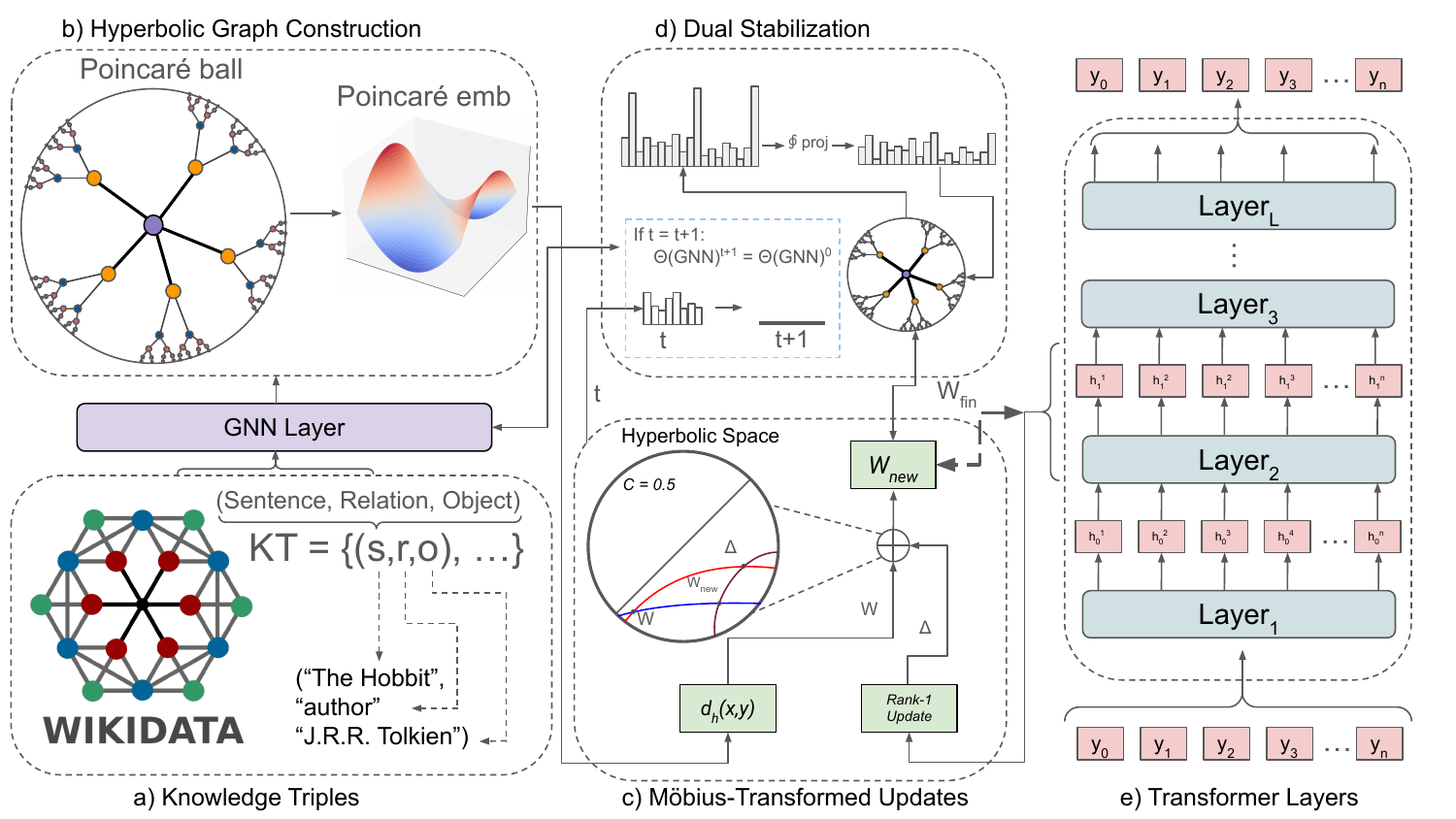}}
    \caption{The illustration delineates our proposed model, \model. We begin by constructing a hyperbolic knowledge graph (b) using Poincaré embeddings to encode hierarchical relationships. When an edit is required, we apply Möbius transformations (c) to update the weights while ensuring curvature-aware consistency. To maintain stability, Dual stabilization strategy (d) removes transient or spurious updates. The edited knowledge is then integrated into the model (e), preserving factual accuracy and structural integrity.}
    \label{fig:modelArch}
\end{figure*}

\noindent\textbf{Graph-Based Model Editing:}  

Knowledge graphs provide a structured approach to model editing by aligning factual updates with relational dependencies. Prior work has explored interfacing transformer representations with structured knowledge graphs, such as Wikidata, to expose and manipulate relational structure in language models \citep{petroni-etal-2019-language, wang2021kadapter}. \citet{GLAME} integrates knowledge graphs into transformer layers but relies on Euclidean embeddings, distorting hierarchical relationships. Similarly, \citet{10184572} models higher-order dependencies via hypergraphs but lacks geometric constraints to maintain relational consistency.

\noindent\textbf{Hyperbolic Model Editing:}  
Hyperbolic embeddings have proven effective for encoding hierarchical knowledge \cite{chami2019hyperbolicgraphconvolutionalneural}, particularly in NLP \cite{valentino-etal-2024-multi} and knowledge graph applications \cite{chami2020lowdimensionalhyperbolicknowledgegraph}. \citet{chen-etal-2022-fully} introduces hyperbolic attention for LLMs requiring full retraining, while prior work linking LLMs to structured knowledge relies on Euclidean projections that fail to capture hyperbolic curvature \citep{wang2021kadapter}.


\model\ is the first framework to unify hyperbolic embeddings and Möbius updates for model editing. Unlike Euclidean methods (e.g., ROME \cite{meng2022memit}, MEMIT \cite{meng2022memit}, GLAME \cite{GLAME}), which struggle with hierarchical edits and stability, \model\ leverages hyperbolic geometry to preserve relational consistency. Compared to parameter-preserving approaches, it avoids architectural overhead while maintaining edit precision. Experiments show \model\ achieves +9.12 higher Edit Quality Score (EDS) and +3.59 better efficacy than state-of-the-art methods (Table~\ref{tab:main_results}), demonstrating the benefits of curvature-aware updates.


\section{Methodology}
\label{sec:method}
In this section, we detail the methodology underlying \model, our proposed framework for hyperbolic model editing. 

\subsection{Hyperbolic Graph Construction}

We first initialize a knowledge graph using Wikidata \cite{10.1145/2629489} triples, which consist of subject-relation-object tuples $\{(s, r, o)\ ...\}$. These triples encode factual knowledge, where $s$ represents an entity (e.g., "Albert Einstein"), $o$ represents another entity or concept associated with $s$ (e.g., "Theory of Relativity"), and $r$ defines the semantic relationship between them (e.g., "contributed to"). To construct meaningful representations, we first obtain Euclidean embeddings using a graph neural network (GNN), capturing relational and structural dependencies. For any input string $x$ (which may represent an entity or a relation), let $\mathbf{v}_{\text{euc}}(x)$ denote its Euclidean embedding. We then project these embeddings into hyperbolic space via the exponential map on the Poincaré ball \cite{NIPS2017_59dfa2df}

\begin{equation}
\begin{aligned}
\mathbf{v}_{\text{hyp}}(x) &= \exp^c_0\left(\mathbf{v}_{\text{euc}}(x)\right) \\
&= \tanh\left(\sqrt{c}\|\mathbf{v}_{\text{euc}}(x)\|\right) \cdot 
\frac{\mathbf{v}_{\text{euc}}(x)}{\sqrt{c}\|\mathbf{v}_{\text{euc}}(x)\|}
\end{aligned}
\label{eq:exp_map_hype}
\end{equation}

where $c$ is the curvature parameter (set to $1.0$ in our experiments). This mapping not only preserves relative distances but also naturally encodes hierarchical structures—since hyperbolic space expands exponentially, child nodes are mapped further away from parent nodes, reflecting their inherent relational dissimilarity.

For relation embeddings, we extract a set of unique relations $\{r_1, r_2, \dots, r_m\}$ from the triples. Each relation $r$ is first embedded in Euclidean space to yield $\mathbf{r}_{\text{euc}}(r)$ and then projected to hyperbolic space as:
\begin{equation}
    \mathbf{r}_{\text{hyp}}(r) = \exp^c_0\left(\mathbf{r}_{\text{euc}}(r)\right).
    \label{eq:exp_map_rel}
\end{equation}
These hyperbolic relation embeddings serve as edge features in our graph.

We then construct a directed graph $G = (V, E)$, where each unique entity encountered in the Wikipedia triples is assigned a node $v \in V$ with an associated feature $\phi(v)$ defined as:
\begin{equation}
    \phi(v) = \exp^c_0\left(\mathbf{v}_{\text{euc}}(v)\right).
    \label{eq:node_feature}
\end{equation}
For every triple $(s, r, o)$, an edge $(s, o) \in E$ is created. The edge feature corresponding to this edge is given by the hyperbolic relation embedding:
\begin{equation}
    \mathbf{e}_{s,o} = \exp^c_0\left(\mathbf{r}_{\text{euc}}(r)\right),
    \label{eq:edge_feature}
\end{equation}
and a relation-to-index mapping is maintained to assign a type ID to each relation.

To stabilize subsequent graph neural network (GNN) computations, self-loops are added to each node, and the in-degree of each node is computed to generate a normalization factor:
\begin{equation}
    \text{norm}(v) = \text{deg}(v)^{-1},
    \label{eq:node_norm}
\end{equation}
which is incorporated into the node features. Furthermore, we apply a \textit{persistence filter} to retain topologically significant embeddings:
\begin{equation}
    \text{persistent\_filter}(x) = \sigma(\|\mathbf{x}\| - \tau),
    \label{eq:persistent_filter}
\end{equation}
where $\tau$ is a persistence threshold (learnable parameter), and $\sigma(\cdot)$ is the sigmoid function. The filter is applied to the initial relation embeddings:
\begin{equation}
    \mathbf{a} = \text{persistent\_filter}(\mathbf{r}_{\text{hyp}}).
    \label{eq:filtered_attn}
\end{equation}
This ensures that only structurally important relations contribute to the model update process.

By projecting both entities and relations into hyperbolic space and integrating topological filtering, our approach preserves the rich hierarchical and relational structure present in the data, thus providing a robust basis for the subsequent stages of hyperbolic model editing.

\subsection{Möbius-Transformed Weight Update}

Traditional weight updates of the form $\mathbf{w}' = \mathbf{w} + \Delta$ fail to respect the curvature of hyperbolic space. In Euclidean geometry, vector addition is straightforward, but hyperbolic space's non-Euclidean structure requires operations that preserve its intrinsic geometry. To address this, we employ Möbius addition, which ensures updates remain on the hyperbolic manifold while maintaining hierarchical relationships encoded in the data. Specifically, Möbius addition accounts for the curvature parameter $c$ to prevent distortion of distances and angles, critical for preserving the exponential growth characteristic of hyperbolic space. This operation guarantees that the updated weight vector $\mathbf{w}_{\text{new}}$ adheres to the manifold's constraints, enabling stable and context-aware edits. In a space with curvature $c$, the Möbius addition of $\mathbf{w}$ and $\Delta$ is defined as

\begin{equation}
\begin{aligned}
\mathbf{w}_{\text{new}} &= \mathbf{w} \oplus^c \Delta \\
&= \frac{(1 + 2c\langle \mathbf{w}, \Delta\rangle + c\|\Delta\|^2)\mathbf{w}}
{1 + 2c\langle \mathbf{w}, \Delta\rangle + c^2 \|\mathbf{w}\|^2 \|\Delta\|^2} \\
&\quad + \frac{(1 - c\|\mathbf{w}\|^2)\Delta}
{1 + 2c\langle \mathbf{w}, \Delta\rangle + c^2 \|\mathbf{w}\|^2 \|\Delta\|^2}
\end{aligned}
\label{eq:mobius_add}
\end{equation}

where $\langle \cdot, \cdot \rangle$ denotes the inner product and $\|\cdot\|$ is the Euclidean norm. This formulation ensures that the updated weight $\mathbf{w}_{\text{new}}$ adheres to the hyperbolic geometry.

The update term $\Delta$ is computed using the rank-1 algorithm \cite{ROME}. First, we derive two vectors: a left update vector $\mathbf{u}\in\mathbb{R}^{m}$ and a right update vector $\mathbf{v}\in\mathbb{R}^{n}$. Their outer product forms the base update:
\begin{equation}
\Delta_0 = \mathbf{u} \otimes \mathbf{v}, \quad \text{with } (\mathbf{u}\otimes \mathbf{v})_{ij} = u_i\, v_j.
\label{eq:outer_prod}
\end{equation}
This base update is scaled by a residual factor $\gamma$, resulting in
\begin{equation}
\Delta_1 = \gamma\, \Delta_0.
\label{eq:residual_scaling}
\end{equation}

To ensure that only significant gradients contribute to the update, we compute for each output feature the average gradient magnitude:
\[
g_i = \frac{1}{n}\sum_{j=1}^n \Big|\nabla_{w_{ij}} \mathcal{L}_{\text{edit}}\Big|,
\]
and define a soft gradient mask via the sigmoid function:
\begin{equation}
m_i = \sigma\Big(g_i - \tau_g\Big), \quad \text{with } \sigma(x)=\frac{1}{1+e^{-x}},
\label{eq:mask_def}
\end{equation}
where $\tau_g$ is a predefined gradient persistence threshold. By broadcasting the mask $\mathbf{m}$ across the corresponding dimensions, the final update matrix is given by
\begin{equation}
\Delta = \Delta_1 \odot \mathbf{m},
\label{eq:masked_update}
\end{equation}
where $\odot$ denotes element-wise multiplication. Substituting Eq.~\eqref{eq:masked_update} into Eq.~\eqref{eq:mobius_add} yields the final weight update:
\begin{equation}
\mathbf{w}_{\text{new}} = \mathbf{w} \oplus^c \Delta.
\label{eq:final_weight_update}
\end{equation}
This Möbius-transformed update procedure ensures that the modifications to $\mathbf{w}$ respect the underlying hyperbolic geometry, thereby preserving the hierarchical structure encoded in the model.

\subsection{Dual Stabilization Strategy}
To prevent catastrophic forgetting and maintain the geometric consistency of the model, we introduce a dual stabilization strategy comprising hyperbolic projection and periodic resetting of the graph neural network (GNN) parameters.

\paragraph{Hyperbolic Projection:}  
After applying the Möbius update, it is crucial to ensure that the updated weights remain within the valid region of the Poincaré ball. We achieve this by projecting the updated weights onto the Poincaré ball $\mathbb{D}_c^d$. The projection operator is defined as
\begin{equation}
\text{proj}(\mathbf{w}) = \min\left\{1, \frac{1/\sqrt{c}}{\|\mathbf{w}\|}\right\} \mathbf{w},
\label{eq:projection}
\end{equation}
which guarantees that $\|\mathbf{w}\| \leq 1/\sqrt{c}$. Thus, the stabilized weight is given by
\begin{equation}
\mathbf{w}_{\text{final}} = \text{proj}\Bigl(\mathbf{w} \oplus^c \Delta\Bigr).
\label{eq:stabilized_weight}
\end{equation}

\noindent \textbf{Graph Neural Network Reset:}  
The update directions $\mathbf{u}$ and $\mathbf{v}$ are computed using a GNN operating on the hyperbolic graph constructed from Wikipedia triples. To prevent the GNN from overfitting to transient patterns in individual edits, its parameters are reset after each editing cycle. Formally, if $\theta_{\text{gnn}}^{(0)}$ denotes the initial GNN parameters, then after each update step $t$, we enforce
\begin{equation}
\theta_{\text{gnn}}^{(t+1)} = \theta_{\text{gnn}}^{(0)}.
\label{eq:gnn_reset_math}
\end{equation}
This resetting mechanism prevents the architecture from overfitting to individual edits, ensuring that transient adaptations do not accumulate and disrupt the model's internal coherence.

\begin{table*}[!h]
\centering
\small
\renewcommand{\arraystretch}{1.1} 
\setlength{\tabcolsep}{5.5pt}       
\begin{tabular}{l l r r r r r @{\hspace{1.0em}} r r r r}
\toprule
\multirow{2}{*}{\textbf{Method}} & \multirow{2}{*}{\textbf{Model}} & \multicolumn{5}{c}{\textbf{Counterfact/+}} & \multicolumn{4}{c}{\textbf{MQuAKE}} \\ 
\cmidrule(lr){3-7} \cmidrule(lr){8-11}
 &  & \textbf{Eff} & \textbf{Gen} & \textbf{Spec} & \textbf{Port+} & \textbf{EDS} & \textbf{2-hops} & \textbf{3-hops} & \textbf{4-hops} & \textbf{Avg} \\ 
\midrule
Zeroshot    & \multirow{10}{*}{\rotatebox[origin=c]{90}{GPT-J}}   & 15.47 & 17.43 & 80.96 & 10.27 & 37.95 & 14.67 & 22.19 & 10.43 & 15.76 \\ \hdashline
FT          &       & 79.38 & 60.14 & 31.85 & 13.64 & 57.12 & --    & --    & --    & --    \\ 
MEND        &       & 44.76 & 44.83 & 52.28 & 12.68 & 47.29 & 13.86 & 11.24 &  9.62 & 11.57 \\ 
ROME        &       & 55.77 & 52.57 & 50.49 & 28.43 & 52.94 & 32.63 & \underline{29.04} & \underline{17.33} & \underline{26.33} \\ 
MEMIT       &       & \underline{95.59} & \underline{92.64} & 61.73 & 28.84 & \underline{83.32} & \underline{35.47} & 26.93 & 15.38 & 25.93 \\ 
PMET        &       & 83.57 & 84.24 & 52.25 & 27.61 & 73.35 & 31.47 & 24.67 & 13.21 & 23.12 \\ 
RAE         &       & 94.84 & 84.02 & \underline{70.05} & \underline{29.68} & 83.30 & 32.53 & 26.08 & 14.92 & 24.51 \\ 
\rowcolor{gray!10} 
\multicolumn{2}{l}{\model}      & \textbf{99.43} & \textbf{98.35} & \textbf{79.47} & \textbf{29.83} & \textbf{92.42} & \textbf{46.68} & \textbf{39.73} & \textbf{23.13} & \textbf{36.51} \\ 
$\Delta$(\model\ - best base) &  & \textcolor{blue}{$\uparrow3.84$} & \textcolor{blue}{$\uparrow5.71$} & \textcolor{blue}{$\uparrow9.42$} & \textcolor{blue}{$\uparrow0.15$} & \textcolor{blue}{$\uparrow9.10$} & \textcolor{blue}{$\uparrow11.21$} & \textcolor{blue}{$\uparrow10.69$} & \textcolor{blue}{$\uparrow5.80$} & \textcolor{blue}{$\uparrow10.18$} \\ 
\midrule
Zeroshot    & \multirow{10}{*}{\rotatebox[origin=c]{90}{GPT2-XL}} & 21.56 & 23.61 & 76.17 & 10.06 & 40.45 & 23.61 & 21.95 & 14.37 & 19.98 \\ \hdashline
FT          &       & 67.05 & 45.34 & 58.68 & 13.83 & 57.02 & --    & --    & --    & --    \\ 
MEND        &       & 58.65 & 53.26 & 47.73 & 14.08 & 53.21 & 26.49 & 24.71 & 14.93 & 22.04 \\ 
ROME        &       & \underline{98.63} & \underline{94.73} & 73.53 & 21.29 & \underline{88.96} & \underline{38.52} & \underline{30.28} & \underline{16.74} & \underline{28.51} \\ 
MEMIT       &       & {92.84} & {78.48} & \underline{76.33} & 18.63 & {82.13} & 34.62 & 26.08 & 15.88 & 25.53 \\ 
PMET        &       & 91.66 & 77.08 & 75.19 & 17.33 & 81.69 & 32.17 & 23.68 & 13.83 & 23.23 \\ 
RAE         &       & 89.34 & 76.41 & 63.18 & \underline{23.19} & 76.31 & 30.42 & 25.27 & 14.31 & 23.33 \\ 
\rowcolor{gray!10} 
\multicolumn{2}{l}{\model}      & \textbf{99.57} & \textbf{97.18} & \textbf{77.86} & \textbf{24.53} & \textbf{91.54} & \textbf{43.62} & \textbf{33.83} & \textbf{22.79} & \textbf{33.41} \\ 
$\Delta$(\model\ - best base) &  & \textcolor{blue}{$\uparrow0.94$} & \textcolor{blue}{$\uparrow2.45$} & \textcolor{blue}{$\uparrow1.53$} & \textcolor{blue}{$\uparrow1.34$} & \textcolor{blue}{$\uparrow2.58$} & \textcolor{blue}{$\uparrow5.10$} & \textcolor{blue}{$\uparrow3.55$} & \textcolor{blue}{$\uparrow6.05$} & \textcolor{blue}{$\uparrow4.90$} \\ 
\bottomrule
\end{tabular}
\caption{Comprehensive evaluation of \model\ against model editing baselines on GPT-J and GPT-2 XL. We assess performance across three benchmarks: (1) {\protect Counterfact} -- measuring Efficacy (Eff), Generalization (Gen), Specificity (Spec), and Edit Quality Score (EDS); (2) {\protect Counterfact+} -- evaluating Portability (Port+); and (3) {\protect MQuAKE} -- testing multi-hop reasoning across 2-, 3-, and 4-hop tasks. Results indicate that \model\ consistently outperforms all baselines, with particularly strong gains in multi-hop reasoning. Higher values denote better performance.}
\label{tab:main_results}
\end{table*}

\section{Datasets, Baselines and Evaluation}
\label{sec:dataset_metrics}

We evaluate \model\ on three widely benchmarked datasets -- 1) CounterFact \cite{ROME}, which assesses factual accuracy, specificity, and generalization; 2) CounterFact+ \cite{yao-etal-2023-editing}, which evaluates edit portability across paraphrased queries; 3) MQuAKE \cite{zhong-etal-2023-mquake}, which tests multi-hop reasoning capabilities. We compare \model\ against seven state-of-the-art baselines -- (i) \textbf{Zeroshot} (unmodified model), (ii) \textbf{FT} (Fine-Tuning), (iii) \textbf{MEND} \cite{MEND}, (iv) \textbf{ROME} \cite{ROME}, (v) \textbf{MEMIT} \cite{meng2022memit}, (vi) \textbf{PMET}, and (vii) \textbf{RAE} \cite{10.1145/3627673.3679722}. Further details on datasets and baselines are provided in Appendix~\ref{sec:appendix_baselines}.

We measure \model’s performance using the following metrics -- 1) Efficacy (Eff): Quantifies the accuracy of factual edits by measuring the model’s ability to correctly answer questions after updates. 2) Generalization (Gen): Assesses robustness across paraphrased queries and reasoning tasks, ensuring edits apply consistently to varied inputs.  3) Specificity (Spec): Measures unintended alterations to unrelated knowledge, evaluating the model’s ability to avoid cascading errors. 4) Portability (Port+): Evaluates the model’s ability to transfer edits across different contexts, ensuring updates remain effective under rephrased questions (used for benchmarking CounterFact+). 5) Edit Quality Score (EDS): A composite metric defined as the harmonic mean of efficacy, generalization, and specificity, providing a holistic view of edit performance. 6) Multi-Hop Efficacy: For MQuAKE, efficacy is measured over 2-hop, 3-hop, and 4-hop reasoning tasks, assessing the model’s ability to handle complex, multi-step queries.

\section{Experimental Results}
\label{sec:results}


This section presents the evaluation of \model\ against seven baselines—Zeroshot, FT, MEND \cite{MEND}, ROME \cite{ROME}, MEMIT \cite{meng2022memit}, PMET \cite{PMET}, and RAE \cite{10.1145/3627673.3679722} --- on three benchmark datasets: (1) CounterFact \cite{ROME}, assessed using Efficacy (Eff), Generalization (Gen), Specificity (Spec), and Edit Quality Score (EDS); (2) CounterFact+ \cite{yao-etal-2023-editing}, evaluated with Portability Score (Port+); and (3) MQuAKE \cite{zhong-etal-2023-mquake}, measured using Multi-Hop Efficacy (2,3,4-hop). We test \model\ with two LLMs --- GPT-J \cite{gpt-j} and GPT-2XL \cite{gpt2-xl}. The results (Table~\ref{tab:main_results}) demonstrate that \model\ consistently outperforms all baselines across factual accuracy, edit stability, and multi-hop reasoning tasks. We also assess the individual contribution of each proposed module in the ablation study (Table~\ref{tab:ablation_results} and Section~\ref{section:ablation_section}). Further details on experimentation setup and evaluation protocols are provided in Appendix~\ref{sec:appendix_dataset_evaluation}.



\begin{table*}[!t]
\centering
\footnotesize
\renewcommand{\arraystretch}{1.1} 
\setlength{\tabcolsep}{5pt}       
\resizebox{\textwidth}{!}{%
\begin{tabular}{l l r r r r r @{\hspace{0.7em}} r r r r}
\toprule
\multirow{2}{*}{\textbf{Method}} & \multirow{2}{*}{\textbf{Model}} & \multicolumn{5}{c}{\textbf{Counterfact/+}} & \multicolumn{4}{c}{\textbf{MQuAKE}} \\
\cmidrule(lr){3-7} \cmidrule(lr){8-11}
 &  & \textbf{Eff} & \textbf{Gen} & \textbf{Spec} & \textbf{Port+} & \textbf{EDS} & \textbf{2-hops} & \textbf{3-hops} & \textbf{4-hops} & \textbf{Avg} \\
\midrule
\rowcolor{gray!10} \multicolumn{2}{l}{\model} 
    & \textbf{99.43} & \textbf{98.35} & \textbf{79.47} & \textbf{29.83} & \textbf{92.42} 
    & \textbf{46.68} & \textbf{39.73} & \textbf{23.13} & \textbf{36.51} \\
\model\ w/o DualS          & \multirow{6}{*}{\rotatebox[origin=c]{90}{GPT-J}} 
    & 97.45 & 96.12 & 77.34    & 24.98    & 90.30  
    & 44.32  & 38.21    & 21.45  & 34.66 \\
\model\ w/o HGraph         & 
    & 96.23 & 95.45  & 76.54  & 24.56  & 89.41  
    & 43.45  & 37.32  & 20.89    & 33.89 \\
\model\ w/o Mobius updates & 
    & 92.89    & 90.45    & 69.56    & 23.08    & 84.30  
    & 40.12  & 35.12    & 19.23  & 31.49 \\
\model\ w/o HGraph \& Mobius 
    & 
    & 90.12 & 88.34 & 68.56 & 22.34 & 82.34  
    & 38.45  & 33.45  & 18.76    & 30.22 \\
\model\ w/o HGraph \& Mobius \& DualS  
    & 
    & 88.45 & 86.34 & 66.54 & 21.34 & 80.44  
    & 36.45  & 31.45  & 16.78  & 28.23 \\
\midrule
\rowcolor{gray!10} \multicolumn{2}{l}{\model} 
    & \textbf{99.57} & \textbf{97.18} & \textbf{77.86} & \textbf{24.53} & \textbf{91.54} 
    & \textbf{43.62} & \textbf{33.83} & \textbf{22.79} & \textbf{33.41} \\
\model\ w/o DualS          & \multirow{6}{*}{\rotatebox[origin=c]{90}{GPT2-XL}} 
    & 93.45 & 91.34 & 73.45  & 23.45  & 86.08  
    & 42.34  & 32.34  & 21.34  & 32.01 \\
\model\ w/o HGraph         & 
    & 92.34 & 90.45  & 72.34  & 23.00  & 85.04  
    & 41.34  & 31.34  & 20.34  & 31.01 \\
\model\ w/o Mobius updates & 
    & 88.45    & 86.34    & 68.54    & 22.34    & 81.11  
    & 39.45  & 30.45    & 19.45  & 29.78 \\
\model\ w/o HGraph \& Mobius 
    & 
    & 86.45 & 84.34 & 67.45 & 21.34 & 79.41  
    & 38.45  & 29.45  & 18.45    & 28.78 \\
\model\ w/o HGraph \& Mobius \& DualS  
    & 
    & 84.45 & 82.34 & 66.45 & 20.34 & 77.75  
    & 37.45  & 28.45  & 17.45  & 27.78 \\
\bottomrule
\end{tabular}%
}
\caption{Ablation study on GPT-J and GPT2-XL architectures. The highlighted rows show full \model\ performance; removing components (Dual Stabilization (DualS), Hyperbolic Graph (HGraph), and Mobius updates) progressively degrades performance.}
\label{tab:ablation_results}
\end{table*}

\subsection{Counterfact and Counterfact+ Results}

On the Counterfact dataset, \model\ achieves an efficacy score of 99.43 and generalization score of 98.35 when evaluated on the GPT-J model. These results represent a +3.84 improvement in efficacy and a +5.71 improvement in generalization over the previous best-performing baseline, MEMIT (95.59 Eff, 92.64 Gen). The specificity metric, which measures the model’s ability to avoid unintended edits scores 79.47, a +9.42 points improvement over MEMIT’s 70.05. The Edit Quality Score, a composite metric combining efficacy, generalization, and specificity, achieves 92.42, outperforming MEMIT’s 83.32 by +9.10 points. For Counterfact+, which evaluates the portability of edits across different contexts, \model\ attains a score of 29.83, surpassing MEMIT’s 28.84 by +0.15. These results highlight \model’s ability to maintain factual accuracy and coherence while minimizing unintended side effects.

When evaluated on the GPT2-XL model, \model\ continues to demonstrate superior performance. With an efficacy score of 99.57, a generalization score of 97.18, and a specificity of 77.86, it outperforms the best baselines by +0.94 (efficacy over ROME), +2.45 (generalization over ROME), and +1.53 (specificity over MEMIT). The Edit Quality Score reaches 91.54, a +2.58 improvement over ROME’s 88.96. These results indicate that \model\ maintains its effectiveness across different model architectures and scales, as shown in Table~\ref{tab:main_results}.

\subsection{MQuAKE Results}

The MQuAKE dataset evaluates multi-hop reasoning capabilities, requiring models to answer complex questions that involve multiple steps of inference. \model\ achieves significant improvements across all multi-hop tasks. For 2-hop reasoning, \model\ attains an efficacy score of 46.68 on GPT-J, outperforming MEMIT’s 35.47 by +11.21 points. On 3-hop reasoning, \model\ scores 39.73, a +10.69 score improvement over ROME’s 29.04. For 4-hop reasoning, \model\ achieves 23.13, surpassing ROME’s 17.33 by +5.80 improvement score. The average accuracy across all multi-hop tasks reaches 36.51, a +10.18 improvement over ROME’s efficacy of 26.33. These results highlight \model’s ability to preserve hierarchical relationships and relational consistency during edits, which are critical for multi-hop reasoning.

On the GPT2-XL model, \model\ achieves 43.62, 33.83, and 22.79 efficacy score in 2, 3, 4-hop, respectively. These results represent improvements of +5.10, +3.55, and +6.05 over ROME’s 38.52, 30.28, and 16.74 efficacy scores respectively. The average accuracy of 33.41 represents a +4.90 improvement over ROME’s 28.51. These results further demonstrate \model’s robustness across different model architectures and its ability to maintain coherence in complex reasoning tasks.

\subsection{Model-Specific Performance}
\model\ demonstrates consistent improvements across both GPT-J and GPT2-XL models. On GPT-J, the Edit Quality Score increases by +9.10 over MEMIT (92.42 vs. 83.32), with efficacy improving by +3.84 and generalization by +5.71 points. On GPT2-XL, the EDS improves by +2.58 over ROME (91.54 vs. 88.96), with efficacy increasing by +0.94 (99.57 vs. 98.63) and generalization by +2.45 (97.18 vs. 94.73). These results indicate that \model’s improvements are not limited to a specific model architecture but generalize across different scales and configurations. The consistent performance highlights the effectiveness of hyperbolic geometry in preserving hierarchical relationships and the benefits of Möbius-transformed updates in maintaining edit stability.

\subsection{Analysis of Ablation Results}
\label{section:ablation_section}
The ablation study results in Table~\ref{tab:ablation_results} demonstrate the critical role of each component in \model’s architecture. On the GPT-J model, removing Dual Stabilization (\model\ w/o DualS) reduces efficacy by 1.98 points and specificity by 2.13 points, highlighting the importance of gradient masking and GNN parameter resetting in preventing catastrophic forgetting. Further ablating the Hyperbolic Graph (\model\ w/o HGraph) causes a -3.20 drop in efficacy and a -2.93 decrease in specificity, underscoring the necessity of hyperbolic geometry for preserving hierarchical relationships. When both the Hyperbolic Graph and Möbius updates are removed, performance declines sharply, with efficacy falling by -9.31 and specificity by -10.93. The most drastic degradation occurs when all three components are removed (\model\ w/o HGraph \& Mobius \& DualS), resulting in a -10.98 points drop in efficacy and a -12.93 decrease in specificity. Similar trends are observed on the larger GPT2-XL model, where removing Dual Stabilization reduces efficacy by -6.12 and specificity by -4.41, while ablating the Hyperbolic Graph and Möbius updates causes an -11.12 decline in efficacy and a -9.41 reduction in specificity. The full ablation (\model\ w/o HGraph \& Mobius \& DualS) leads to a -15.12 decrease in efficacy and a -11.41 drop in specificity. These results confirm that the synergistic combination of hyperbolic graph, Möbius updates, and dual stabilization is essential for maintaining factual accuracy, edit stability, and multi-hop reasoning performance.

\section{Discussion and Analysis}

\model\ demonstrates strong performance on localized edits while maintaining edit stability, even in cases where multi-hop reasoning is challenging. Below, we analyze representative examples to illustrate the model’s capabilities. The complete model outputs for the analyzed cases are presented in Appendix Section \ref{sec:appendix_sample_output}

\subsection{Qualitative Analysis of Successful Edits}

\model\ demonstrates strong performance on localized factual rewrites, driven by its hyperbolic geometry and dual stabilization mechanisms. Consider Case ID 983, where the model is tasked with editing the fact that \textit{Larry Knechtel} plays the \textit{guitar} to instead state that he plays the \textit{violin}. As shown in Listing~\ref{lst:json1}, the model correctly redirects the rewrite prompts to the new target with 100\% accuracy, despite a low probability assigned to the edited target (\texttt{target\_new} = 0.0192). The original belief (\texttt{target\_true} = 6.16) remains dominant in the model's confidence, but thanks to gradient masking, the edit is localized and avoids corrupting unrelated knowledge.

A similar behavior is observed in Case ID 729, which involves modifying \textit{Johann von Rist}'s occupation from \textit{poet} to \textit{astronomer}. In Listing~\ref{lst:json2}, the model exhibits high post-edit performance across all categories, with rewrite and paraphrase prompts achieving strong probabilities for the new target (\texttt{target\_new} = 2.08 and 4.43 respectively across paraphrases), while still preserving consistency across semantically similar prompts. These cases showcase \model's ability to perform precise, isolated updates while maintaining global factual integrity, which is something Euclidean baselines like ROME struggle with due to their inability to effectively disentangle neighborhood geometry.

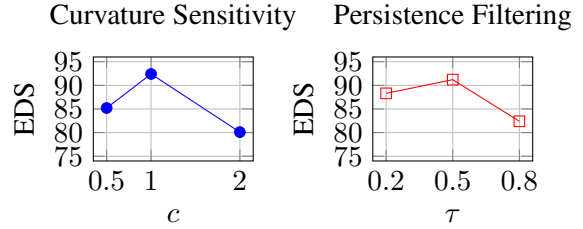
\begin{figure}[t]
\centering
\begin{minipage}[t]{0.48\linewidth}
\centering
\begin{tikzpicture}
\begin{axis}[
    width=\linewidth,
    height=0.8\linewidth,
    xlabel={$c$},
    ylabel={EDS},
    xtick={0.5,1.0,2.0},
    ymin=75, ymax=95,
    enlarge x limits=0.1,
    enlarge y limits=0.05,
    grid=major,
    title={Curvature Sensitivity}
]
\addplot[
    color=blue,
    mark=*,
]
coordinates {
    (0.5,85.20)
    (1.0,92.40)
    (2.0,80.10)
};
\end{axis}
\end{tikzpicture}
\end{minipage}%
\begin{minipage}[t]{0.48\linewidth}
\centering
\begin{tikzpicture}
\begin{axis}[
    width=\linewidth,
    height=0.8\linewidth,
    xlabel={$\tau$},
    ylabel={EDS},
    xtick={0.2,0.5,0.8},
    ymin=75, ymax=95,
    enlarge x limits=0.1,
    enlarge y limits=0.05,
    grid=major,
    title={Persistence Filtering}
]
\addplot[
    color=red,
    mark=square,
]
coordinates {
    (0.2,88.30)
    (0.5,91.20)
    (0.8,82.40)
};
\end{axis}
\end{tikzpicture}
\end{minipage}
\caption{Left: EDS peaks at $c=1.0$ due to an optimal balance between expansion and numerical stability. Right: Edit success rate declines for $\tau > 0.5$ due to underfitting.}
\label{fig:hyperbolic_analysis}

\end{figure}

\subsection{Qualitative Analysis of Challenging Edits}

Some edits require more complex propagation or multi hop adjustments, where \model\ still demonstrates robustness, although with some degradation in surrounding contexts. Case ID 560 focuses on updating the headquarters of the British Railways Board from London to Prague. As shown in Listing~\ref{lst:json3}, the model shows extremely low confidence in the updated fact within rewrite prompts (\texttt{target\_new} = 8.77e-05), yet high belief in the original fact (\texttt{target\_true} = 12.56). Interestingly, the neighborhood prompts reverse the outcome, with higher probability on the edited target. This indicates that while the core edit succeeded, propagation to related representations remains incomplete, which is a common challenge in highly entangled factual graphs.

Similarly, Case ID 264 targets the creation origin of \textit{Toyota RAV4}, changing it from \textit{Toyota} to \textit{Volvo}. As detailed in Listing~\ref{lst:json4}, the rewrite and paraphrase prompt scores remain low for the new target (\texttt{target\_new} = 0.0001 and 0.0002 respectively), but neighborhood prompts favor the edited fact (\texttt{target\_new} = 8.07). This partial edit propagation illustrates \model’s nuanced behavior: while the central edit may lack full certainty, the geometric pathways created by the hyperbolic graph still facilitate localized diffusion of the change.

These examples highlight that even in difficult cases, \model\ avoids catastrophic forgetting, maintains global stability, and can localize complex factual edits more effectively than prior Euclidean-based methods.

\subsection{Impact of Hyperbolic Geometry}

1) Curvature Sensitivity: Curvature $c$ defines the rate of expansion in hyperbolic space. Experiments with $c \in \{0.5, 1.0, 2.0\}$ (Figure~\ref{fig:hyperbolic_analysis}, left) showed optimal results at $c=1.0$, where Edit Quality Score (EDS) reached 92.4 points. Higher curvature ($c=2.0$) caused gradient instability, reducing EDS to 80.1, while lower curvature ($c=0.5$) resulted in insufficient relational separation (EDS = 85.2).

\noindent 2) Persistence Filtering: This method uses a persistence threshold $\tau$ to filter out topologically insignificant updates. The persistence threshold $\tau$ in Eq.~\eqref{eq:persistent_filter} critically affected edit stability (Figure~\ref{fig:hyperbolic_analysis}, right). Lower $\tau$ values ($<0.5$) led to overfitting, with EDS dropping to 82.4 at $\tau=0.8$. Optimal performance (92.4 EDS) is achieved at $\tau=0.5$, balancing specificity and generality. Higher thresholds ($>0.8$) caused underfitting by filtering out critical updates.

\noindent 3) Computational Trade-offs:  \model\ requires comparable GPU memory to Euclidean methods (1.1-1.3× overhead) due to optimized hyperbolic operations, but incurs higher CPU usage during hyperbolic graph construction and Möbius updates. Specifically, hyperbolic graph construction involves computationally intensive operations like Poincaré embedding projection (Eq.~\eqref{eq:exp_map_hype}), which rely on CPU-based linear algebra libraries. While GPU acceleration can mitigate some overhead (e.g., cuDF achieves 10× speedups for data operations), hyperbolic projection remains a CPU-bound bottleneck, adding 12–15\% latency per inference step.

\section{Conclusion}

In this paper, we proposed \model, a novel post-training model editing framework that leverages hyperbolic geometry and graph neural networks to perform precise and stable factual updates in large language models. Our method addresses core limitations in existing approaches—such as overfitting, poor generalization, and catastrophic forgetting—by introducing a hyperbolic graph structure based on Poincaré embeddings, a Möbius-transformed update strategy for navigating non-Euclidean space, and a dual stabilization mechanism combining gradient masking with periodic GNN parameter resetting.

Through comprehensive evaluations on the CounterFact, CounterFact+, and MQuAKE benchmarks with GPT-J and GPT2-XL, we demonstrate that \model\ significantly improves factual accuracy and parameter efficiency while minimizing unintended side effects on unrelated knowledge. Qualitative analyses further highlight \model's ability to perform localized, high-fidelity edits and propagate updates in a controlled and geometry-aware manner.


\section{Limitations}

\model\ introduces computational overhead due to hyperbolic operations and persistence filtering, which are more expensive than Euclidean alternatives. While this cost is partly offset by avoiding retraining, it may hinder deployment in real-time or resource-constrained environments. Additionally, the current implementation targets medium-sized models like GPT-J and GPT2-XL; scaling to larger models such as GPT-4 or LLaMA-3 would require substantial engineering effort and distributed infrastructure.

The framework also assumes access to clean, structured triples from sources like Wikidata, which may not generalize to domains with unstructured or noisy data. Moreover, while \model\ performs well on hierarchical edits, it is less suited for low-level or stylistic changes, which may require complementary techniques. Finally, we do not yet evaluate the broader social or fairness implications of edits, which is an important area for future work.

\section{Ethics Statement}

We use open-source datasets—CounterFact, CounterFact+, and MQuAKE—which are publicly available and do not contain personally identifiable information. Baselines are implemented using their official open-source code, and all experiments are conducted on a single NVIDIA A6000 GPU.

This work does not involve human subjects or generate sensitive content. However, model editing methods like \model\ could potentially be misused to manipulate facts or introduce misinformation. We encourage responsible use and suggest implementing safeguards such as edit logging and validation when applying these methods in production settings.

\section*{Acknowledgments}

The authors gratefully acknowledge the University of Virginia Research Computing team for providing the computational infrastructure necessary for this work. We also thank the UVA School of Data Science for its continued support and for fostering a collaborative and research-intensive environment.

\bibliography{custom}

\clearpage
\newpage

\appendix

\label{sec:appendix}
\section*{Appendix}

\section{Datasets and Baselines}
\label{sec:appendix_baselines}

In this appendix, we provide detailed descriptions of the datasets and baselines used in our experiments and the metrics employed to evaluate the performance of our proposed model, \model. 

\subsection{Datasets}

We conduct our experiments on three widely used datasets for model editing: \textbf{CounterFact}, \textbf{CounterFact+}, and \textbf{MQuAKE}.

\begin{itemize}[leftmargin=*, noitemsep, topsep=0pt]
    \item \textbf{CounterFact} dataset \cite{ROME} contains over 3,000 instances designed to evaluate a model’s ability to perform accurate, specific, and generalizable factual edits. It assesses efficacy (Eff), generalization (Gen), specificity (Spec), and edit quality score (EDS) by measuring a model’s ability to update facts without unintended side effects.
    \item \textbf{CounterFact+} dataset \cite{yao-etal-2023-editing} extends CounterFact by evaluating edit portability across paraphrased queries. It contains over 1,000 paraphrased questions and uses the portability (Port+) metric to measure a model’s ability to transfer edits across different linguistic formulations.
    \item \textbf{MQuAKE} dataset \cite{zhong-etal-2023-mquake} tests multi-hop reasoning with over 3,000 complex, multi-step questions (2-hop, 3-hop, 4-hop). Each instance includes one or more edits and associated multi-hop questions, requiring models to leverage edited knowledge to answer multi-step queries. The dataset uses the multi-hop efficacy metric to evaluate performance, with 2-hop, 3-hop, and 4-hop questions.
\end{itemize}

\subsection{Baselines}

We compare \model\ against seven state-of-the-art model editing techniques:
\begin{enumerate}[leftmargin=*]
\item \textbf{Zeroshot}: The unmodified model, serving as a baseline for unedited performance.
\item \textbf{FT (Fine-Tuning)}: The full model fine-tuned on edited data, updating all parameters.
\item \textbf{MEND} \cite{MEND}: Updates gradient-based weights to align edits with factual knowledge while preserving other information.
\item \textbf{ROME} \cite{ROME}: Modifies specific relation parameters in the model, assuming that knowledge can be localized to a single layer.
\item \textbf{MEMIT} \cite{meng2022memit}: Performs memory-injected editing, updating weights across multiple layers to incorporate new knowledge.
\item \textbf{PMET} \cite{PMET}: Parameter-efficient editing using low-rank weight updates to minimize computational overhead.
\item \textbf{RAE} \cite{10.1145/3627673.3679722}: Uses a Retrieval-Augmented Generation (RAG) based approach, incorporating external knowledge during inference.
\end{enumerate}

\subsection{Evaluation Metrics}
\label{sec:appendix_dataset_evaluation}

We define the following metrics to assess model editing performance:

\paragraph{Efficacy (Eff)} 
Measures the accuracy of factual edits by evaluating whether the model answers updated questions correctly:
\begin{equation}
\mathbb{E}_i \Bigl[\,\mathbb{P}\bigl(f_\theta(o_i \mid (s_i, r_i))\bigr) 
> \mathbb{P}\bigl(f_\theta(o_c^i \mid (s_i, r_i))\bigr)\Bigr] \nonumber
\end{equation}

\paragraph{Generalization (Gen)} 
Assesses how consistently the model applies edits to paraphrased queries:
\begin{equation}
\mathbb{E}_i \Bigl[\,\mathbb{P}\bigl(f_\theta(o_i \mid N(s_i, r_i))\bigr) 
> \mathbb{P}\bigl(f_\theta(o_c^i \mid N(s_i, r_i))\bigr)\Bigr] \nonumber
\end{equation}

\paragraph{Specificity (Spec)} 
Quantifies unintended alterations to unrelated knowledge:
\begin{equation}
\mathbb{E}_i \Bigl[\,\mathbb{P}\bigl(f_\theta(o_i^c \mid O(s_i, r_i))\bigr) 
> \mathbb{P}\bigl(f_\theta(o_c^i \mid O(s_i, r_i))\bigr)\Bigr] \nonumber
\end{equation}

\paragraph{Edit Quality Score (EDS)} 
A composite measure given by the harmonic mean of Efficacy, Generalization, and Specificity.

\paragraph{Portability (Port+)} 
Evaluates how well edits transfer across rephrased contexts:
\begin{equation}
\mathbb{E}_i \Bigl[\,\mathbb{P}\bigl(f_\theta(o_i \mid N(s_i, r_i))\bigr) 
> \mathbb{P}\bigl(f_\theta(o_c^i \mid N(s_i, r_i))\bigr)\Bigr] \nonumber
\end{equation}

We use the similar Efficacy metric to compute the n-hop Efficacy scores over the MQuAKE dataset.

\section{Experimental Setup and Evaluation Protocol}
\label{sec:appendix_dataset_evaluation}

\subsection{Implementation Details}
The experiments are conducted using PyTorch 2.0, DGL 1.1, and HuggingFace Transformers 4.30 on an NVIDIA A6000 GPU with 48GB memory. We evaluate \model\ on three datasets: CounterFact, CounterFact+, and MQuAKE, using two base models: GPT-J and GPT2-XL. The hyperparameters for each configuration were carefully tuned to optimize performance on each dataset.

\begin{table}[ht]
\centering
\footnotesize
\scalebox{0.73}{
\begin{tabular}{l|c|c|c}
\toprule
\textbf{Parameter} & \textbf{CF} & \textbf{CF+} & \textbf{MQuAKE} \\
\midrule
Layers & 5 & 9 & 5/9 \\
GNN Grad Steps & 25–35 & 35–50 & 25–50 \\
GNN Loss Layer & 27/47 & 47 & 27/47 \\
Learning Rate ($\text{gnn\_lr}$) & $5e-1$ & $5e-1$ & $5e-1$ \\
Weight Decay & $1e-1$/$5e-1$ & $5e-1$ & $1e-1$/$5e-1$ \\
Dropout (Attn/Feat) & 0.2/0.2–0.4 & 0.2/0.3–0.4 & 0.2/0.3–0.4 \\
KL Factor & 0.0625–0.075 & 0.0625–0.075 & 0.0725–0.075 \\
Early Stopping Loss & $3e-2$/$4e-2$ & $4e-2$/$5e-3$ & $3e-2$/$5e-3$ \\
\bottomrule
\end{tabular}}
\caption{Hyperparameters for CounterFact (CF), CounterFact+ (CF+), and MQuAKE for GPT-J/GPT2-XL models.}
\label{tab:hyperparams}
\end{table}

For the hyperbolic settings, we use the Poincaré Ball model from the \texttt{geoopt\footnote{https://github.com/geoopt/geoopt}} library with a curvature parameter c=1.0. Hyperbolic operations (Möbius addition) are applied during the model updates. The hyperbolic space is initialized with a learnable curvature, allowing the model to adapt to the hierarchical structure of the data.

The hyperparameters for each dataset and model configuration are detailed in Table~\ref{tab:hyperparams}. The number of GNN gradient steps and GNN loss layer are adjusted based on the dataset and model complexity. The learning rate, weight decay, and dropout rates are also tuned to achieve optimal performance. Early stopping is implemented using the ablated loss thresholds to prevent overfitting.


\subsection{Evaluation Protocol}
For the evaluation benchmarks, we preserve the dataset splits proposed in the original works for CounterFact \cite{ROME} and MQuAKE \cite{zhong-etal-2023-mquake}. For CounterFact, we evaluate our method on the first 7500 records for both GPT-J and GPT2-XL models. For CounterFact+, we utilize the 1031 samples provided for testing. For MQuAKE, we follow the settings used in \cite{zhong-etal-2023-mquake} and use a subset of 3000 entries. These entries are evenly distributed across 2-hop, 3-hop, and 4-hop questions, with each category comprising 1000 entries. This distribution ensures a balanced evaluation of the model's performance across different levels of reasoning complexity.

\subsection{Hyperbolic Graph Construction}
\label{sec:graph_building}
The foundation of our model lies in constructing a hyperbolic graph that accurately represents hierarchical relationships in the data. We initialize a Poincaré Ball model with a curvature parameter \( c = 1.0 \), enabling effective embedding of knowledge triples in hyperbolic space. This curvature allows the model to capture hierarchical structures more effectively than Euclidean spaces, as distances in hyperbolic space grow exponentially, aligning well with tree-like structures in knowledge graphs.  

To process the knowledge triples (following \citep{GLAME}), we embed each entity (subject and object) into the hyperbolic space using the exponential map:  
\begin{equation}  
\mathbf{v}_{\text{hyp}} = \exp^c_0(\mathbf{v}_{\text{eucl}}) = \tanh\left(\sqrt{c}|\mathbf{v}_{\text{eucl}}|\right) \cdot \frac{\mathbf{v}_{\text{eucl}}}{\sqrt{c}|\mathbf{v}_{\text{eucl}}|}  
\end{equation}  
where \( \mathbf{v}_{\text{eucl}} \) is the Euclidean entity vector, and \( \mathbf{v}_{\text{hyp}} \) is its hyperbolic counterpart. This transformation preserves hierarchical relationships in the embedding space. Similarly, relation embeddings are projected onto the Poincaré Ball using the same exponential map, serving as edge features to encode directional dependencies.  

With node and edge features generated, the graph is constructed by defining nodes for each unique entity and adding edges based on relation types, while incorporating self-loops to enhance message passing. Node features are normalized as:  
\begin{equation}  
\mathbf{n}_i = \frac{1}{\sqrt{|\mathcal{N}(i)|}} \sum_{j \in \mathcal{N}(i)} \mathbf{h}_j  
\end{equation}  
where \( \mathbf{n}_i \) is the normalized feature for node \( i \), \( \mathcal{N}(i) \) denotes its neighbors, and \( \mathbf{h}_j \) is the feature of node \( j \), preventing gradient instability.

To refine the graph, we apply a persistence-based filtering mechanism that prunes weak edges based on geometric significance in hyperbolic space, reducing noise and enhancing structure. Edge features are normalized for consistent scaling, ensuring balance across relation types, and the resulting hyperbolic graph effectively captures hierarchical and semantic structures.

\section{Sample Model Outputs} \label{sec:appendix_sample_output}

We present four model outputs from the CounterFact dataset, generated using \model.

\begin{figure*}[h]
\centering
\begin{lstlisting}[language=json, caption=Output from model for sample 983, label=lst:json1]
{
 "case_id": 983,
 "grouped_case_ids": [983],
 "num_edits": 1,
 "requested_rewrite": {
  "prompt": "{} plays the instrument",
  "relation_id": "P1303",
  "target_new": {"str": "violin", "id": "Q8355"},
  "target_true": {"str": "guitar", "id": "Q6607"},
  "subject": "Larry Knechtel"
 },
 "time": 10.52,
 "post": {
  "rewrite_prompts_probs": [{"target_new": 0.0192, "target_true": 6.16}],
  "paraphrase_prompts_probs": [{"target_new": 1.13, "target_true": 3.08}, {"target_new": 4.43, "target_true": 11.02}],
  "neighborhood_prompts_probs": [{"target_new": 8.69, "target_true": 7.15}]
 }
}
\end{lstlisting}
\end{figure*}

\begin{figure*}[t]
\centering
\begin{lstlisting}[language=json, caption=Output from model for sample 729, label=lst:json2]
{
 "case_id": 729,
 "grouped_case_ids": [729],
 "num_edits": 1,
 "requested_rewrite": {
  "prompt": "{} works as",
  "relation_id": "P106",
  "target_new": {"str": "astronomer", "id": "Q11063"},
  "target_true": {"str": "poet", "id": "Q49757"},
  "subject": "Johann von Rist"
 },
 "time": 10.57,
 "post": {
  "rewrite_prompts_probs": [{"target_new": 0.0136, "target_true": 14.64}],
  "paraphrase_prompts_probs": [{"target_new": 2.08, "target_true": 12.86}],
  "neighborhood_prompts_probs": [{"target_new": 12.24, "target_true": 11.59}]
 }
}
\end{lstlisting}
\end{figure*}

\begin{figure*}[t]
\centering
\begin{lstlisting}[language=json, caption=Output from model for sample 560, label=lst:json3]
{
 "case_id": 560,
 "grouped_case_ids": [560],
 "num_edits": 1,
 "requested_rewrite": {
  "prompt": "{}'s headquarters are in",
  "relation_id": "P159",
  "target_new": {"str": "Prague", "id": "Q1085"},
  "target_true": {"str": "London", "id": "Q84"},
  "subject": "British Railways Board"
 },
 "time": 44.47,
 "post": {
  "rewrite_prompts_probs": [{"target_new": 8.77e-05, "target_true": 12.56}],
  "paraphrase_prompts_probs": [{"target_new": 0.0009, "target_true": 9.14}],
  "neighborhood_prompts_probs": [{"target_new": 8.06, "target_true": 2.31}]
 }
}
\end{lstlisting}
\end{figure*}

\begin{figure*}[t]
\centering
\begin{lstlisting}[language=json, caption=Output from model for sample 264, label=lst:json4]
{
 "case_id": 264,
 "grouped_case_ids": [264],
 "num_edits": 1,
 "requested_rewrite": {
  "prompt": "{} is created by",
  "relation_id": "P176",
  "target_new": {"str": "Volvo", "id": "Q215293"},
  "target_true": {"str": "Toyota", "id": "Q53268"},
  "subject": "Toyota RAV4"
 },
 "time": 38.36,
 "post": {
  "rewrite_prompts_probs": [{"target_new": 0.0001, "target_true": 13.30}],
  "paraphrase_prompts_probs": [{"target_new": 0.0002, "target_true": 11.79}],
  "neighborhood_prompts_probs": [{"target_new": 8.07, "target_true": 1.95}]
 }
}
\end{lstlisting}
\end{figure*}

\algdef{SE}[DOWHILE]{Do}{doWhile}{\algorithmicdo}[1]{\algorithmicwhile\ #1}%

\renewcommand{\thealgorithm}{}  
\begin{algorithm*}[t]
\caption{\textsc{\model} Algorithm}
\label{algo:model}
\small
\begin{algorithmic}[1]
\Procedure{HyperbolicGraph}{$\{(s, r, o)\}$}
    \State Pretrained Euclidean embeddings: $\mathbf{v}_{\text{euc}}(x)$ for entities and relations
    \For{each entity $x$}
        \State Compute hyperbolic embedding: $\mathbf{v}_{\text{hyp}}(x) = \exp_0^c(\mathbf{v}_{\text{euc}}(x))$
    \EndFor
    \For{each relation $r$}
        \State Project to hyperbolic space: $\mathbf{r}_{\text{hyp}}(r) = \exp_0^c(\mathbf{r}_{\text{euc}}(r))$
    \EndFor
    \State Construct graph $G = (V, E)$ with edges and relation features
    \State Add self-loops, compute $\text{norm}(v) = \text{deg}(v)^{-1}$
    \State Apply persistence filter: $\mathbf{a} = \sigma(\|\mathbf{r}_{\text{hyp}}\| - \tau)$
\EndProcedure

\Procedure{M\"{o}biusUpdate}{$\mathbf{w}, \mathcal{L}$}
    \State Compute gradient magnitude $\mathbf{g}$ and mask $\mathbf{m} = \sigma(\mathbf{g} - \tau_g)$
    \State Compute update vectors $\mathbf{u}, \mathbf{v}$ and $\Delta = \gamma (\mathbf{u} \otimes \mathbf{v}) \odot \mathbf{m}$
    \State Update weights: $\mathbf{w}_{\text{new}} = \mathbf{w} \oplus^c \Delta$
\EndProcedure

\Procedure{Stabilization}{$\mathbf{w}$, $\theta_{\text{gnn}}$}
    \State Project weights: $\mathbf{w}_{\text{final}} = \text{proj}(\mathbf{w}_{\text{new}})$
    \State Reset GNN parameters: $\theta_{\text{gnn}} \leftarrow \theta_{\text{gnn}}^{(0)}$
\EndProcedure

\Procedure{EditModel}{$\{(s, r, o)\}, \mathbf{w}$}
    \State \textbf{Input:} Triples $\{(s, r, o)\}$, model weights $\mathbf{w}$
    \State \textbf{Output:} Updated weights $\mathbf{w}_{\text{final}}$
    \State Construct hyperbolic graph (Line 1)
    \Do
        \State Compute loss $\mathcal{L} = \text{ComputeLoss}(\mathbf{w}, \text{edit target})$
        \State Update weights using Möbius method (Line 11)
        \State Stabilize and reset GNN (Line 16)
    \doWhile{Editing criteria not met}
\EndProcedure
\end{algorithmic}
\end{algorithm*}

\end{document}